\documentclass[lettersize,journal]{IEEEtran}
\usepackage{amsmath,amsfonts}
\usepackage{amssymb}
\usepackage{algorithmic}
\usepackage{algorithm}
\usepackage{array}
\usepackage[caption=false,font=normalsize,labelfont=sf,textfont=sf]{subfig}
\usepackage{textcomp}
\usepackage{stfloats}
\usepackage{url}
\usepackage{verbatim}
\usepackage{graphicx}
\usepackage{cite}
\usepackage{caption}
\usepackage{amsmath}
\usepackage{multirow}
\usepackage{makecell}
\usepackage{booktabs}
\usepackage[table]{xcolor}

\begin{document}

\title{SeMv-3D: Towards Concurrency of Semantic and Multi-view Consistency in General Text-to-3D Generation}

\author{Xiao Cai$^{1}$,
        Pengpeng Zeng$^{2}$,
	Lianli~Gao$^{1}$,
        Sitong Su$^{1}$,
        Heng~Tao~Shen$^{1, 2}$, and Jingkuan~Song$^{2}$,
 \\
        $^{1}$University of Electronic Science and Technology of China\\
        $^{2}$Tongji University
}



\maketitle


\begin{abstract}
General Text-to-3D (GT23D) generation is crucial for creating diverse 3D content across objects and scenes, yet it faces two key challenges: 1) ensuring semantic consistency between input text and generated 3D models, and 2) maintaining multi-view consistency across different perspectives within 3D. Existing approaches typically address only one of these challenges, often leading to suboptimal results in semantic fidelity and structural coherence. To overcome these limitations, we propose \textbf{\textit{SeMv-3D}}, a novel framework that jointly enhances semantic alignment and multi-view consistency in GT23D generation. At its core, we introduce \textbf{Triplane Prior Learning (TPL)}, which effectively learns triplane priors by capturing spatial correspondences across three orthogonal planes using a dedicated Orthogonal Attention mechanism, thereby ensuring geometric consistency across viewpoints. Additionally, we present \textbf{Prior-based Semantic Aligning in Triplanes (SAT)}, which enables consistent any-view synthesis by leveraging attention-based feature alignment to reinforce the correspondence between textual semantics and triplane representations. Extensive experiments demonstrate that our method sets a new state-of-the-art in multi-view consistency, while maintaining competitive performance in semantic consistency compared to methods focused solely on semantic alignment. These results emphasize the remarkable ability of our approach to effectively balance and excel in both dimensions, establishing a new benchmark in the field.
\end{abstract}


\begin{figure*}[!htb]
    \centering
    \includegraphics[width=1\linewidth]{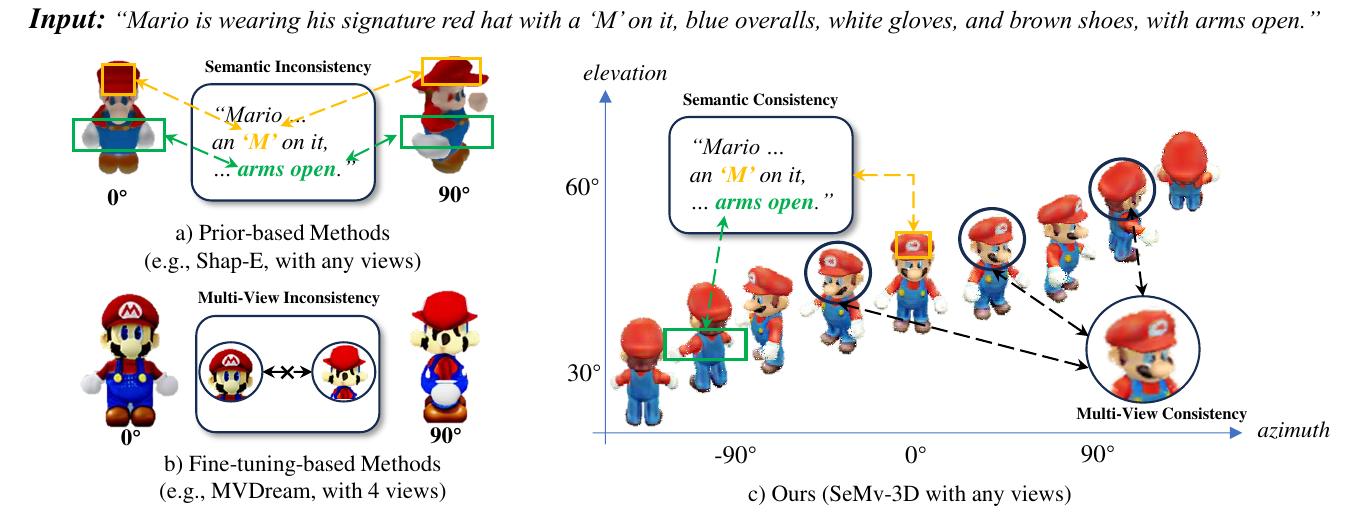}
     \caption{\textbf{Comparison with SOTA baselines and our SeMv-3D.} The two mainstream lines of general text-to-3d: a) Prior-based methods and b) Fine-tuning-based methods struggle with two core challenges: semantic inconsistency and multi-view inconsistency, respectively. Our SeMv-3D c) can ensure the concurrency of semantic and multi-view consistency.}
    \label{fig:intro}
\end{figure*}

\section{Introduction}

\IEEEPARstart{T}{ext}-to-3D generation (T23D) aims to generate corresponding 3D content based on text prompts with a broad range of applications, including games, movies, virtual/augmented reality, and robotics. Previous works mainly focus on a per-scene optimization problem \cite{DreamFusion, Magic3D, ProlificDreamer, Fantasia3D}, which yields fine texture and geometric details. 
However, these methods lack the ability for general generation, as each new prompt requires retraining to approximate the textual semantics.
To overcome this issue, general text-to-3D (GT23D) has been proposed by learning a generic model capable of synthesizing various objects in a feed-forward manner. General text-to-3D faces two core challenges: a) \textbf{Semantic Consistency}, which requires semantic alignment of the generated 3D context with the text, and b) \textbf{Multi-view Consistency}, which maintains coherence across multiple 3D views. 

Benefiting from the great breakthroughs in the text-to-image diffusion (T2I) models, two research lines of rationalization have recently emerged in general text-to-3D: prior-based methods and fine-tuning-based methods.
Specifically, prior-based methods primarily leverage the T2I models as semantic-visual initialization and subsequently train on large-scale 3D datasets. They solely focus on regressing the corresponding 3D shapes, which naturally ensures consistency across multiple views, such as Shap-E~\cite{Shap-E} and VolumeDiffusion~\cite{VolumeDiffusion}. However, it sacrifices portions of the well-learned semantic alignment information of the original T2I model, inevitably resulting in inconsistency between the generated visuals and their corresponding semantics, presented in Figure \ref{fig:intro}a. Even the most advanced method, Shap-E, fails to fully match the semantics of different components in the prompt.

Conversely, fine-tuning-based methods seek to transfer the strong single-view generation capabilities of pretrained T2I models (e.g., great semantic alignment between text and vision) directly to generate multiple views, such as MVDream~\cite{MVDream} and DreamView~\cite{DreamView}. Yet these methods are inherently ambiguous without explicit 3D constraints, leading to notorious multi-view inconsistency (e.g., multi-face Janus problem, shown as Figure \ref{fig:intro}b) and limited-view.  Thus, how to effectively and simultaneously achieve semantic and multi-view consistency remains to be explored for the general text-to-3D task.

Toward the above goal, we propose a novel framework, named \textit{\textbf{SeMv-3D}}, which learns a consistent triplane prior to ensure uniformity across all views of an object and align its semantics with the text.
Empirically, the triplane has been validated as an effective  3D representation for object modeling~\cite{EG3D, 3DTopia}. Unlike existing methods that directly learn the entire triplane features, we emphasize spatial correspondence within the triplane to capture the underlying 3D details. 
Specifically, we propose a \textbf{T}riplane \textbf{P}rior \textbf{L}earning (\textbf{TPL}) that integrates 3D spatial features into a triplane prior. In practice, TPL first eliminates irrelevant backgrounds or components to preserve essential 3D information and then captures spatial correspondence within triplane space to enhance its visual coherence by our designed Orthogonal Attention. Moreover, we design Prior-based \textbf{S}emantic \textbf{A}ligning in \textbf{T}riplanes (\textbf{SAT}) that deeply interacts between textual and visual features within triplane priors through attention-based alignment, significantly improving semantic consistency. From Figure \ref{fig:intro}c, we can see that our method performs better in multi-view and semantic consistency than other compared methods. 

To summarize, our main contributions are threefold: 
\begin{itemize}

 \item We devise a \textit{\textbf{SeMv-3D}}, a novel general text-to-3D framework, which ensures the concurrency of semantic and multi-view consistency. 

 \item We propose a \textbf{TPL}, which learns a triplane prior to effectively capture consistent 3D features across views. Moreover, we devise a \textbf{SAT} that deeply explores the alignment between textual and 3D visual information, substantially improving semantic consistency.

 \item Extensive experiments demonstrate the superiority of \textit{\textbf{SeMv-3D}} in both multi-view and semantic consistency, achieving state-of-the-art multi-view alignment while maintaining competitive semantic performance.
\end{itemize}

\section{Related works}
\label{sec:related work}
While earlier 3D generation methods focused on reconstructing geometry from single-view~\cite{Zero-1-to-3, wander3d, tang2023make, shuang2025direct3d, tip-singleview, tip-point}, multi-view~\cite{yu2021pixelnerf, wang2021ibrnet, tip-multiview, tip-multiview2}, or other visual inputs such as sketches or point clouds~\cite{mikaeili2023sked, diffusion-point, liu2024sketchdream, tip-sketch}, generating 3D content directly from textual descriptions—commonly referred to as Text-to-3D (T23D)—marks a substantial shift in paradigm. 
T23D aims to synthesize 3D representations (3D voxels, point clouds, multi-view images, and meshes) from textual descriptions. Early works of T23D~\cite{michel2022text2mesh, sella2023vox, vahdat2022lion, mittal2022autosdf, cheng2023sdfusion} directly train generation models on small-scale 3D datasets, which restricted the semantic diversity and geometry fidelity of the 3D outputs.
With the emergence of pretrained Text-to-Image (T2I) diffusion models, recent works utilize semantic-visual prior knowledge of these T2I models for fine-grained and diverse 3D generation. Existing works can be grouped into two categories based on generalization ability: 1) Per-scene Text-to-3D and 2) General Text-to-3D.

\textbf{Per-scene Text-to-3D.}
Per-scene Text-to-3D requires per-scene optimization when generating a new scene. The mainstream idea is using knowledge from pre-trained T2I models to guide the optimization of 3D representations.
DreamFusion~\cite{DreamFusion} employs a technique known as Score Distillation Sampling (SDS). This approach utilizes large-scale image diffusion models~\cite{LDM,Imagen} to iteratively refine 3D models to match specific prompts or images. Similarly, ProlificDreamer~\cite{ProlificDreamer} develops Variational Score Distillation (VSD), a structured variational framework that effectively reduces the over-saturation problems found in SDS while also increasing diversity. Further enhancements are offered by several studies~\cite{Magic123,RichDreamer,ImageDream}, which address the challenges of multiple faces by using diffusion models fine-tuningd on 3D data. The strategy of amortized score distillation is examined in other references~\cite{ATT3D,AToM}. Numerous additional works~\cite{Fantasia3D,Magic3D,TextMesh,HiFA} have substantially improved both the speed and quality of these approaches. 
Despite fine-grained texture details through optimization, these methods usually require a lengthy period, ranging from minutes to hours, to generate only a single object. Contrastly, our approach employs a feed-forward method that requires no per-scene optimization.

\textbf{General Text-to-3D.} 
Methods in General Text-to-3D achieve open-domain T23D without needing additional optimization for each new scene. These methods can be divided into two categories based on their implementation process: fine-tuning-based and prior-based approaches. Prior work SDFusion~\cite{SDFusion} takes dense SDF grids as the 3D representation, which is computational cost and unable to render textures. Point-E~\cite{Point-E} and Shap-E~\cite{Shap-E}, trained on millions of 3D assets, generate point clouds and meshes respectively. 3DGen~\cite{3DGen} combines a triplane VAE for learning latent representations of textured meshes with a conditional diffusion model for generating the triplane features. VolumeDiffusion~\cite{VolumeDiffusion} trains an efficient volumetric encoder to produce training data for the diffusion model. With insufficient 3D data to learn, recent works tend to utilize 2D priors to help the training. Inspired by image-to-3D models~\cite{Zero-1-to-3,One-2-3-45}, image diffusion models are adopted for 3D generation. MVDream~\cite{MVDream} and DreamView~\cite{DreamView} attempt to jointly train the image generation model with high-quality normal images and limited multi-view object images to produce various object images. 
Recently, SPAD~\cite{SPAD} builds upon MVDream to achieve arbitrary view generation. Despite these advancements, current methods still struggle to generate both semantic and multi-view consistent views. In contrast, our approach learns a complete 3D prior, enabling arbitrary view generation while maintaining consistent results across different views.

\section{Methodology}
\begin{figure*}[t]
\centering
\includegraphics[width=\textwidth]{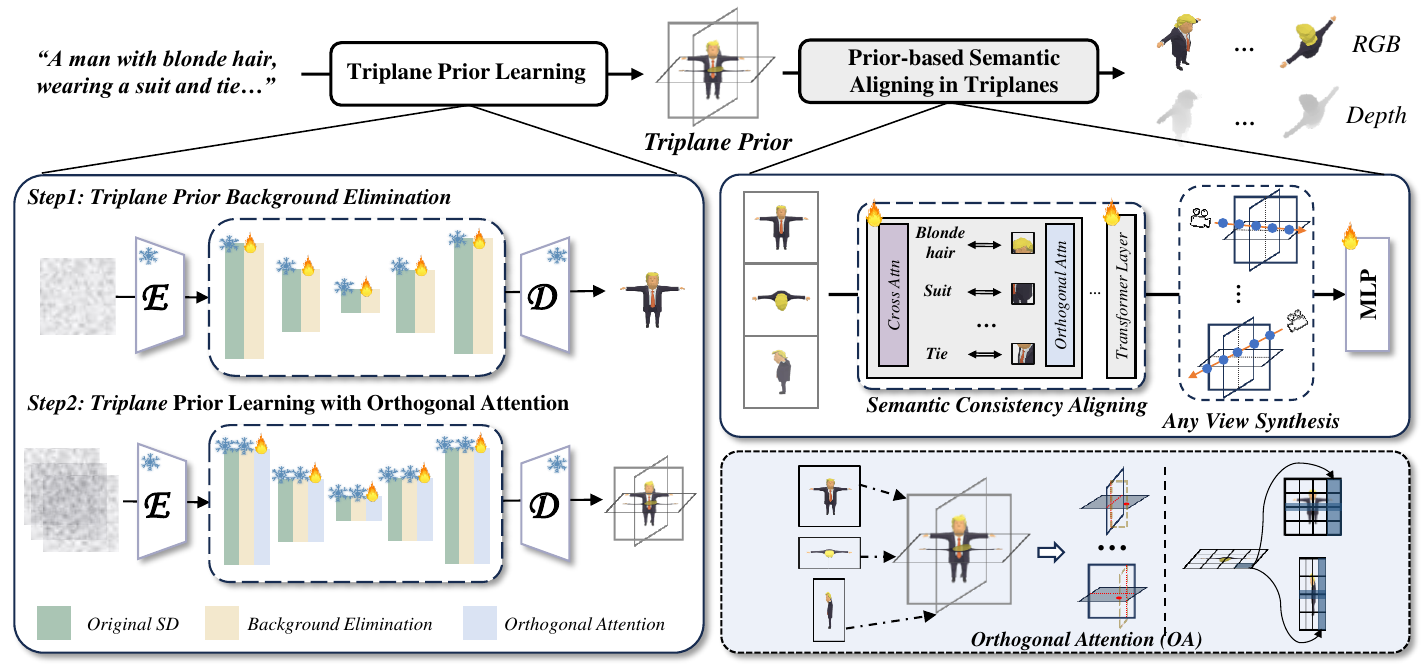}
\caption{\textbf{The overall framework of SeMv-3D.} SeMv-3D consists of two components: 1) Triplane Prior Learning (\textbf{TPL}) that learns a triplane prior to capture consistent 3D visual details and 2) Prior-based Semantic Aligning in Triplanes (\textbf{SAT}) that enhances the alignment between the semantic with 3D content and enables single-step generation of arbitrary views. Here, Orthogonal Attention (\textbf{OA}) focuses on the orthogonal correspondences within the triplane, maintaining triplane consistency.}
\label{fig:pipeline}
\end{figure*}

In this section, we present a triplane prior based GT23D approach \textbf{SeMv-3D}, which comprises two core stages: Triplane Prior Learning (TPL) and Prior-based Semantic Aligning in Triplanes (SAT), as shown in Figure~\ref{fig:pipeline}. 

\subsection{Triplane Prior Learning} 

\subsubsection{\textbf{Problem Formulation}}
Triplane Prior Learning (TPL) is a two-step trained latent diffusion-based~\cite{LDM} model that takes textual descriptions $\mathcal{T}$ as input and generates an explicit triplane prior:  
$P_{\text{tri}} = TPL(\mathcal{T}),$
where \( P_{\text{tri}} = \{ P_{xy}, P_{xz}, P_{yz} \} \) represents the three orthogonal planes (i.e., front, top, and side). This prior preserves spatial correspondence consistency across the three planes, ensuring structural coherence in 3D representation learning.

Within our Triplane Prior Learning (TPL), shown in Figure~\ref{fig:pipeline}, we first conduct Background Elimination (BE) to mitigate the adverse effects of inconsistent backgrounds on the prior learning process. Then, we design an Orthogonal Attention to help learn the spatial correspondence across three planes. 
\label{sec: tpl}
\subsubsection{\textbf{Triplane Prior Background Elimination}}
\label{OR}

To eliminate the adverse impact of highly varied backgrounds across planes, we introduce Background Elimination (BE) to filter out irrelevant elements, facilitating consistent triplane alignment for subsequent steps.

In practice, one residual block and one attention block are plugged into each level of the U-Net network before upscale and downscale, while all other pre-trained layers are frozen during training. The learning objective function can be described as follows:
\vspace{-4pt}
\begin{equation}
\small
\begin{aligned}
    \mathcal{L}_{\text{BE}}=\mathbb{E}_{t \sim [1, T], \mathbf{x}_0, \boldsymbol{\epsilon}_t} 
    \Big[\|\boldsymbol{\epsilon}_t - \\
    \boldsymbol{\epsilon}_{\theta_{\text{BE}}}&(\sqrt{\bar{\alpha}_t}\mathbf{x}^{i}_0 + \sqrt{1 - \bar{\alpha}_t}\boldsymbol{\epsilon}_t, t, \mathcal{T})\|^2 \Big],
\end{aligned}
\end{equation}
where ${\epsilon}_t$ is the added noise for diffusion process for the timestep $t$ on the condition text prompt $\mathcal{T}$, $\theta_{BE}$ is the learnable parameters in BE, $\bar{\alpha}_t$ is the pre-defined hyper-parameters for the sampling scheduler, and $\mathbf{x}^{i}_0$ is a clean object image sampling from random viewpoints.

\subsubsection{\textbf{Triplane Prior Learning with Orthogonal Attention}}
\label{TO}

To effectively learn the triplane priors, we capture and align the spatial correspondences across the planes using our dedicated Orthogonal Attention (OA), facilitating the delivery of consistent and reliable 3D prior knowledge.
Specifically, the OA focuses on the orthogonal spatial relationship between three planes and correlates the orthogonality to ensure consistency.  

For example, as shown in the bottom-right corner of Figure~\ref{fig:pipeline}, given a pixel $(a, b, -)$ in the latent $xy$-plane $\textbf{P}_{xy}$ which needs to focus on pixels in the other two orthogonal planes,  it should intersect with all pixels sharing the same x-coordinate ($a$) in the $xz$-plane $\textbf{P}_{xz}$ and all pixels with the same y-axis coordinate ($b$) in the $yz$-plane $\textbf{P}_{yz}$, as well as the pixels on the cross-line between the corresponding planes. The Orthogonal Attention can be expressed as follows:
\begin{equation}
    \begin{aligned}
        \text{OA}(\textbf{P}_{xy}|\textbf{P}_{xz},\textbf{P}_{yz})&=\text{OA}_{x}(\textbf{P}_{xy},\textbf{P}_{xz})+\text{OA}_{y}(\textbf{P}_{xy},\textbf{P}_{yz}),\\
        \text{OA}(\textbf{P}_{xz}|\textbf{P}_{xy},\textbf{P}_{yz})&=\text{OA}_{x}(\textbf{P}_{xz},\textbf{P}_{xy})+\text{OA}_{z}(\textbf{P}_{xz},\textbf{P}_{yz}),\\
        \text{OA}(\textbf{P}_{yz}|\textbf{P}_{xz},\textbf{P}_{xy})&=\text{OA}_{z}(\textbf{P}_{yz},\textbf{P}_{xz})+\text{OA}_{y}(\textbf{P}_{yz},\textbf{P}_{xy}),
    \end{aligned}
\end{equation}
and
\begin{equation}
\small
\begin{aligned}
    \text{OA}_{i}(\textbf{P}_{1},\textbf{P}_{2}) = \prod_{M \in \textbf{P}_{1}}& \text{softmax}\left(\frac{W_Q(M)W_K(N)^{\intercal}}{\sqrt{d_{W_K(N)}}}\right)W_V(N), \\
    & \ \text{s.t.i} \ \ i \in \{x,y,z\},
\end{aligned}
\end{equation}

where
\begin{equation}
\small
    \begin{aligned}
    M =  \{&K| K  \in \textbf{P}_{1}\}, 
    N =  \{K| K \in \textbf{P}_{2} \ \& \\
    &(Coord_{i}(M)=Coord_{i}(K) \  | \ K \in (\textbf{P}_{1} \cap \textbf{P}_{2})) \},
    \end{aligned}
\end{equation}

where $\textbf{$P_i$}$ represents the i-th plane, $W_Q$, $W_K$, and $W_V$ refer to query, key, and value mapping functions, and $Coord_{i}(\cdot)$ indicates the i-axis coordinate. 

Then, in the triplane prior learning process, we append a Triplane Learning (TL) module incorporating OA after each BE module. We freeze all other components and only optimize added TL modules with the triplane supervision, whose objective function can be expressed as follows:
\begin{equation}
\small
    \begin{aligned}
    \mathcal{L}_{\text{TL}}=\mathbb{E}_{t \sim [1, T], \mathbf{x}_0, \boldsymbol{\epsilon}_t} & \sum_{i\in\{xy,xz,yz\}} \\
    & \Big[ \|\boldsymbol{\epsilon}^i_t -
     \boldsymbol{\epsilon}_{\theta_{\text{TL}}}(\sqrt{\bar{\alpha}_t}\mathbf{x}^{i}_0 + \sqrt{1 - \bar{\alpha}_t}\boldsymbol{\epsilon}^i_t, t, \mathcal{T})\|^2 \Big],
    \end{aligned}
\end{equation}
where $\theta_{TL}$ is the learnable parameters in TL, and $\mathbf{x}^{i}_0$ refers to the front, top, and side views images.

\subsection{Prior-based Semantic Aligning in Triplanes}
\label{sec: svs}
Given the learned consistent triplane prior through our TPL, we introduce Prior-based Semantic Aligning in Triplanes (SAT) to establish fine-grained correspondences between textual and triplane features, enhancing semantic consistency while enabling any-view synthesis in a single feed-forward inference step.

\subsubsection{\textbf{Semantic Consistency Aligning}}
To maintain semantic consistency during triplane construction, we leverage attention-based feature alignment to bind the correspondence between semantic and triplane features.

\noindent\textbf{Text-Triplane Semantic Consistency via Cross-Attention.}
\label{TLT}
We first introduce a Cross-Attention (CA) mechanism to align semantic information with triplane features in global. The CA mechanism effectively integrates the textual semantics with the implicit triplane fields, ensuring that each plane's features are consistently guided by the corresponding semantic context, thereby reinforcing the overall semantic coherence in the generated 3D content.
Specifically, we extract semantic features $feat_{text}$ from condition text prompt $T$ and triplane features $feat_{tri}$ from triplane prior $P_{tri}$. These features are then integrated into the global semantic-aligned feature space $feat_{tri}^G$ using cross-attention, denoted as:
\begin{equation}
    feat_{tri}^{G} = CA(feat_{tri}, feat_{text}).
\end{equation}

\noindent\textbf{Triplane Semantic Consistency via Orthogonal Attention.}
To further strengthen fine-grained semantic consistency, we reintroduce Orthogonal Attention (OA), which takes into account the inherent structural properties of the triplane, ensuring more accurate alignment at the local level.
In practice, we enable the integrated features to adaptively align local visual region features with their corresponding semantics through Orthogonal Attention (OA), as represented by:
\begin{equation}
    feat_{tri}^{G\&L}=OA(feat_{tri}^{G}).
\end{equation}
During the semantic consistency alignment, we transfer the processed features to the implicit radiance field by a classic transformer model, $Trans(\cdot)$, obtaining triplane latents $f_{Tri}$ that contain ample semantics and 3D information:
\begin{equation}
\begin{aligned}
    f_{Tri} = Trans(feat_{tri}^{G\&L})
\end{aligned}
\end{equation}
\subsubsection{\textbf{Any-View Synthesis}}
\label{strategy}
With the well-learned triplane latents, we design a simple yet effective batch sampling and rendering strategy to enable the generation of any views in a single feed-forward step. 
Following the \cite{EG3D, NeRF}, we employ the triplane latents $f_{Tri}$ as implicit fields for ray sampling and rendering.

\begin{figure*}[t]
\centering
\includegraphics[width=\textwidth]{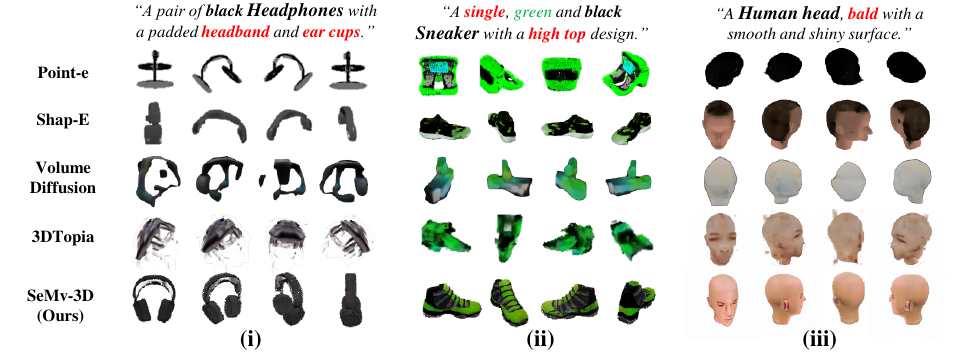}
\caption{\textbf{Qualitative Comparison with Prior-based GT23D Methods.} It shows our method maintains better semantic and multi-view consistency than Prior-based Methods. More results are presented in the \textbf{Suppl. C}}.
\label{fig:qualitative_result_sota_prior}
\end{figure*}
In ray sampling, given a batch of camera positions \(\textbf{o}\), we construct a batch of pixels from different views by tracing ray paths \(\textbf{r}(t) = \textbf{o}_{i} + t\textbf{d}\) along their respective directions \(\textbf{d}\), where $t$ represents the distance along the ray.
Then, for each ray, we sample multiple points along its path, with the sampling range constrained by a near bound $t_n$ and a far bound $t_f$. 
In ray rendering, for each sampled point \(\textbf{r}(t)\) along the ray, we project it onto the three orthogonal triplanes: the \(xy\)-plane, the \(xz\)-plane, and the \(yz\)-plane. The features retrieved from these three projected locations are then concatenated  to obtain the feature representation \(\textbf{f}(\textbf{r}(t))\) for the corresponding sampled point. Typically, for projected points with non-integer coordinates, we interpolate their features from the four nearest pixels to obtain precise representations. Finally, we aggregate the features of all sampled points along each ray to compute the rendered pixels in a batch.
Specifically, we learn two MLP functions (i.e., $S(\cdot)$ and $C(\cdot)$) to predict the density $\sigma$ and color $\textbf{c}$ of each point, as follows:
\begin{equation}
\begin{aligned}
    \sigma(\textbf{r}(t))=S(\textbf{r}(t),\textbf{f}(\textbf{r}(t))), \\ 
    \textbf{c}(\textbf{r}(t)) = C(\textbf{r}(t),\textbf{f}(\textbf{r}(t))), \\ 
\end{aligned}
\end{equation}
Then, we calculate the pixel information, accumulating all sampled points as follows:
\begin{equation}
    \begin{aligned}
    \textbf{Pix}_{\text{rgb}} = \int_{t_n}^{t_f} T(t) \cdot \sigma(\textbf{r}(t)) \cdot \textbf{c}(\textbf{r}(t)) \ dt, 
    \end{aligned}
\end{equation}
where
\begin{equation}
    T(t)=exp(-\int_{t_n}^t \sigma(\textbf{r}(s)) \ ds).
\end{equation}
Generally, RGB pixels can be totally discretely rendered for optimization since they are independent of each other. Once the pixel colors $\textbf{Pix}_{\text{rgb}}$ for a batch are computed, we can obtain the batch images $\textbf{I}$. Similarly, the corresponding masks $\textbf{S}$ and depths $\textbf{D}$ are also derived.
The object function can be expressed as follows:
\begin{equation}
    \begin{aligned}
    \mathcal{L}_{\text{Render}}=\sum^{N}_{i=1} &(\Vert \textbf{I}^i - \textbf{I}^i_\text{GT} \Vert_2 + \lambda_{S} \Vert \textbf{S}^i - \textbf{S}^i_\text{GT} \Vert_2 +\\ 
    &\lambda_{D} \Vert \textbf{D}^i - \textbf{D}^i_\text{GT} \Vert_2 +
    \lambda_{lpips} (\mathcal{L}_{\text{lpips}}(\textbf{I}^i, \textbf{I}^i_\text{GT})),
    \label{loss_render}
    \end{aligned}
\end{equation}
where $N$ indicates the view number used for training, and $\textbf{I}_\text{GT}$, $\textbf{S}_\text{GT}$ and $\textbf{D}_\text{GT}$ refer respectively to the ground truth in pixel, mask and depth. $\mathcal{L}_{\text{lpips}}$~\cite{lpips} is the perceptual loss for better optimization. We set $\lambda_{S}=0.5$, $\lambda_{D}=1$, $\lambda_{lpips}=2$ to balance each item.


\section{Experiments}
In this section, we first conduct comprehensive experiments to demonstrate SeMv-3D's superior multi-view consistency and competitive semantic consistency, comparing it to prior-based GT23D methods in Sec.\ref{sec:comparison_pr} and fine-tuning-based GT23D methods in Sec.\ref{sec:comparison_ft}.  Then, we present ablation studies to analyze the effectiveness of SeMv-3D in Sec.~\ref{sec:ablation}. 

\subsection{Experiment Setup}
\label{sec:exper_details}

\noindent\textbf{Training dataset.}
SeMv-3D is trained on the Objaverse~\cite{Objaverse} dataset. Each object is rendered into 64 multi-view images at a resolution of $512 \times 512$ using PyVista~\cite{pyvista}. Textual descriptions are generated using LLaVA-1.6~\cite{LLaVA16}, resulting in paired text–3D data for pretraining. During the Triplane Prior Learning (TPL) stage, three canonical views (i.e., front, side, and top) are used for supervision. In the Prior-based Semantic Aligning in Triplane (SAT) stage, four views are randomly sampled from the 64 rendered images at each training step to serve as supervision.

\begin{table}[t]
    \centering
    \Large
    \renewcommand{\arraystretch}{1.2}
    \caption{Quantitative Comparison with Prior-based methods.}
    \resizebox{1.0\linewidth}{!}{
    \begin{tabular}{c|ccccc}
    \hline 
     & Point-E & Shap-E & VolumeDiff & 3DTopia & SeMv-3D \\
    \hline
    CLIP Score & 23.43 & 28.90 & 23.51 & 25.87 & \cellcolor{green!10} \textbf{30.26} \\
    Mv-Score  & 86.28/100 & 88.69/100 & 83.80/100 & 84.25/100 & \cellcolor{green!10} \textbf{91.63/100} \\
    \hline
    \end{tabular}
    }
    \label{tab:objective_comparison_prior}
\end{table}

\begin{table}[t]
    \centering
    \Large
    \renewcommand{\arraystretch}{1.2}
    \caption{User Study of SeMv-3D and Prior-based methods.}
    \resizebox{1.0\linewidth}{!}{
    \begin{tabular}{c|ccccc}
    \hline 
     & Point-E & Shap-E & VolumeDiff & 3DTopia & SeMv-3D \\
    \hline
    User Prefer & 7.4\%  & 28.4\%  & 6.1\%  & 14.2\%  & \cellcolor{pink!20} \textbf{43.9\%} \\
    Se Consis   & 3.4\%  & 29.7\%  & 3.4\%  & 12.2\%  & \cellcolor{pink!20} \textbf{51.4\%} \\
    Mv Consis   & 17.6\%  & 24.3\% & 14.2\%  & 11.5\%  & \cellcolor{pink!20} \textbf{32.4\%} \\
    \hline
    \end{tabular}
    }
    \label{tab:subjective_comparison_prior}
\end{table}

\noindent\textbf{Implementation details.} 
We initialize the Triplane Prior Learning (TPL) with Stable Diffusion 2.1~\cite{latent-diffusion}. The Background Elimination (BE) module is trained for 150k steps with a learning rate of \ $5 \times 10^{-4}$, while the Triplane Learning (TL) module is trained for 60k steps with a learning rate of \ $5 \times 10^{-5}$.
The Prior-based Semantic Aligning in Triplane (SAT) is trained for 100k steps with a learning rate of \ $5 \times 10^{-4}$. All training processes are conducted on 8 NVIDIA A6000 GPUs, adopting the AdamW~\cite{AdamW} optimizer for all stages with $\beta_1=0.9$, $\beta_2=0.95$, and weight decay $=0.03$.

\begin{figure*}[t]
\centering
\includegraphics[width=\textwidth]{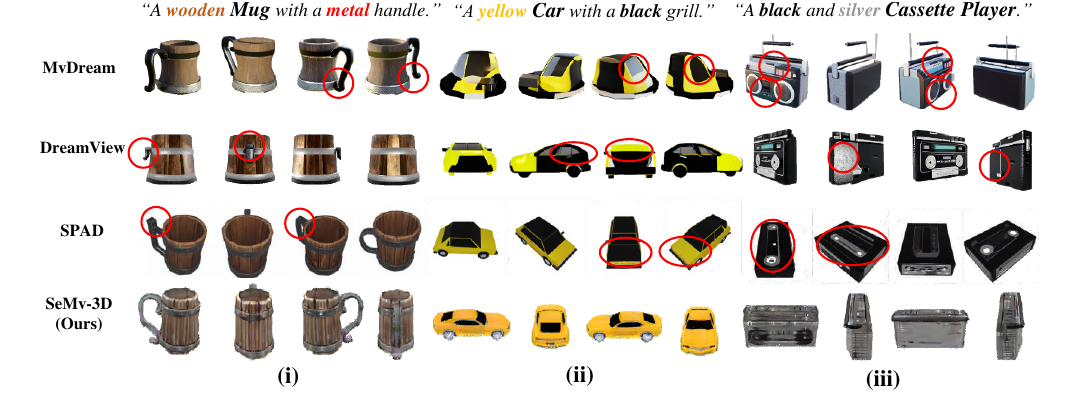}
\caption{\textbf{Qualitative Comparison with Fine-tuning-based GT23D Methods.} It demonstrates that our method achieves superior multi-view consistency compared to fine-tuning-based methods while maintaining comparable semantic consistency. More results are presented in the \textbf{Suppl. C}.}
\label{fig:qualitative_result_sota_ft}
\end{figure*}
\noindent{\textbf{Evaluation Metrics}}
To comprehensively evaluate the performance of text-to-3D generation methods in terms of semantic alignment and multi-view coherence, we adopt two complementary evaluation metrics:
\textbf{1) CLIP Score}~\cite{CLIPScore}: This metric measures the semantic alignment between the input text and the generated 3D object. Specifically, we compute the average CLIP similarity between the input prompt and multiple rendered views of the generated 3D shape. A higher CLIP Score indicates better correspondence between textual semantics and visual appearance, serving as an effective proxy for semantic consistency.
\textbf{2) Multi-view Consistency Score (Mv-Score)}: As there is currently no standardized metric for evaluating multi-view consistency in 3D generation, we adopt GPT-4o~\cite{openai2023gpt4} as an evaluation tool to assess the coherence of 3D objects across multiple viewpoints. For fair comparison, all methods are evaluated using four canonical views—0°, 90°, 180°, and 270°. The Mv-Score comprises four sub-dimensions with a total of 100 points:
(\romannumeral1) \textit{Geometric Consistency (20 pts)}: Assesses the consistency of shape and structural components across different views.
(\romannumeral2) \textit{Texture Consistency (20 pts)}: Evaluates the consistency of textures and surface appearances across views.
(\romannumeral3) \textit{Angular Consistency (20 pts)}: Measures the plausibility of spatial transitions between adjacent viewpoints.
(\romannumeral4) \textit{Overall Consistency (40 pts)}: Provides a holistic judgment of global consistency across all views, integrating geometry, texture, and viewpoint alignment.
More details are provided in \textbf{Suppl. A}.

\noindent\textbf{User Study.} 
We conduct user studies where participants evaluate the results from three perspectives: 1) User Preference (Users Prefer), indicating overall user preference for the generated views; 2) Semantic Consistency (Se consis): measuring how well the generated objects align with the text, similar to CLIP score; and 3) Multi-view Consistency (Mv consis): assessing the consistency of objects across different views.  More details are provided in \textbf{Suppl. B}.

\subsection{Comparison with Prior-based Methods.}
\label{sec:comparison_pr}
To confirm that our method achieves both strong semantic and multi-view consistency, we compare it with state-of-the-art prior-based methods including: Point-E~\cite{Point-E}, Shap-E~\cite{Shap-E}, VolumeDiffusion (abbreviated as VolumeDiff)~\cite{VolumeDiffusion} and 3DTopia~\cite{3DTopia}.

\noindent\textbf{Qualitative Comparisons.}
To provide a more intuitive comparison, we conduct a visual quality evaluation, as shown in Figure \ref{fig:qualitative_result_sota_prior}, 
As we can see, our approach achieves superior semantic and multi-view consistency. In comparison, methods like 3DTopia fail to generate objects that align with the overall semantics, such as the `headphones' in (i). Furthermore, when handling fine-grained semantic details, as shown in (ii) and (iii), these methods generally struggle to capture attributes like `high-top' or `bald,' leading to significant semantic inconsistencies.
In addition, irregular artifacts (such as the VolumeDiffusion results in (i)) appearing across different views exacerbate multi-view inconsistency.

\noindent\textbf{Quantitative Comparison.}
\label{sec.quantitative}
To systematically evaluate our approach, we perform quantitative comparisons against prior-based methods, as shown in Table \ref{tab:objective_comparison_prior}. The results indicate that (\romannumeral1) our method surpasses all prior-based approaches, achieving a SOTA CLIP Score of 30.26, highlighting its superior semantic consistency and generation quality.
(\romannumeral2) Our approach achieves the highest multi-view consistency score, demonstrating fewer artifacts compared to other prior-based methods, leading to more consistent multi-view results. 
The quantitative results above further highlight our significant advantage in both semantic consistency and multi-view consistency

\begin{figure*}
    \centering
    \includegraphics[width=1\linewidth]{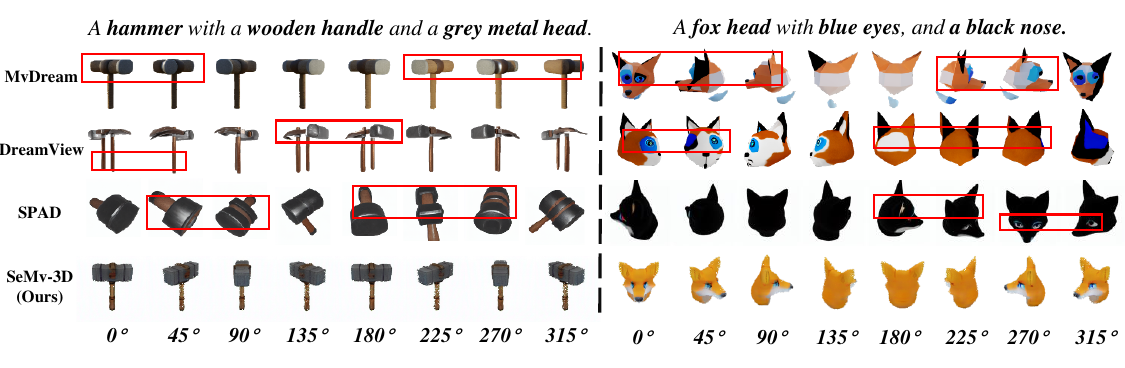}
    \caption{\textbf{Multi-view Consistency Challenges in Fine-tuning-based Methods}: Generating more than four views leads to severe geometric and texture inconsistencies. In contrast, our method demonstrates superior multi-view consistency compared to existing fine-tuning-based methods in any-view synthesis.}
    \label{fig:mov_finetune}
\end{figure*}
\noindent\textbf{User Study.} 
To validate the strength of our method from human perception, we also conduct a user study. 
As shown in the Table \ref{tab:subjective_comparison_prior}, our SeMv-3D is preferred by an average of 43.9\% of users, surpassing all baseline methods. Additionally, it achieves the highest user preference scores for semantic consistency (51.4\%) and view consistency (32.4\%). These results further demonstrate the effectiveness of our approach in maintaining both semantic and multi-view consistency.

\subsection{Comparsion with Fine-tuning-based Methods.}
\label{sec:comparison_ft}
To further demonstrate the effectiveness of SeMv-3D, we compare it against leading fine-tuning-based approaches, including MVDream~\cite{MVDream}, DreamView~\cite{DreamView}, and SPAD~\cite{SPAD}.

\noindent\textbf{Qualitative Comparisons.} 
Similarly, we present a visual comparison with state-of-the-art fine-tuning-based methods in Figure \ref{fig:qualitative_result_sota_ft}.
For symmetrical objects such as the \textit{“Mug”} and \textit{“Car”} (Figure \ref{fig:qualitative_result_sota_ft} (i) and (ii)), fine-tuning-based methods maintain consistency in primary structures but struggle with localized details like handles and windows. Moreover, for texture-rich objects such as the \textit{“Cassette Player”} (Figure \ref{fig:qualitative_result_sota_ft} (iii)), methods like MVDream and DreamView fail to preserve texture details across views, while SPAD produces highly inconsistent results. In comparison, our method maintains excellent consistency in both object structure and texture alignment.
Additionally, \textbf{generating more than 4 views exacerbates inconsistencies in fine-tuning-based methods}, as illustrated in Figure~\ref{fig:mov_finetune}. As the number of views increases, these methods exhibit increasing inconsistencies in both geometry and appearance across viewpoints. In terms of geometry, object structures often change abruptly with viewpoint rotation. For example, the head of the `\textit{hammer}' becomes visibly deformed, and the handle orientation becomes unstable across different angles. In terms of appearance, textures and surface materials are inconsistently rendered. In the case of the `\textit{fox}', the facial region shows noticeable variations in color and shading, and components such as the eyes are occasionally missing or distorted. In contrast, our SeMv-3D produces coherent and stable results across all views, consistently preserving both structural integrity and appearance fidelity in any-view synthesis.

\noindent\textbf{Quantitative Comparisons.}
We also conduct quantitative comparisons between baselines and our method SeMv-3D in Table \ref{tab:objective_comparison_ft}. From the table, we can find that: (\romannumeral1) Our method achieves state-of-the-art performance in multi-view consistency. 
(\romannumeral2) Although our method does not achieve the SOTA CLIP Score, it secures an impressive second place with a score of 30.26, surpassing most fine-tuning-based approaches such as MVDream and SPAD. The higher score of DreamView may result from its emphasis on optimizing 4 limited views, which strengthens semantic consistency but could reduce multi-view consistency, whereas our method ensures consistency across any views.
(\romannumeral3) It demonstrates that, while maintaining superior multi-view consistency, our approach also achieves semantic consistency comparable to state-of-the-art fine-tuning methods, showcasing its strong capability in both aspects.

\noindent\textbf{User Study.}
We conduct a user study to further validate the effectiveness of our method. As shown in Table \ref{tab:subjective_comparison_ft}, the majority of users prefer our results across three aspects. 
Human preferences suggest that users favor results that maintain both multi-view and semantic consistency, confirming that our method effectively balances these two aspects.

\subsection{Ablation Studies}
\label{sec:ablation}
In this section, we conduct comprehensive ablation studies to validate the effectiveness of each component in SeMv-3D, including Triplane Prior Learning (TPL) and Prior-based Semantic Aligning in Triplanes (SAT).

\subsubsection{\textbf{Effectiveness of the Designed Components in TPL}}
Our proposed Triplane Prior Learning (TPL) framework extends a base model through the integration of three key modules: Background Elimination (BE), Triplane Learning (TL), and Orthogonal Attention (OA). To evaluate the contribution and effectiveness of each component, we adopt an incremental ablation strategy—starting from the base model and sequentially adding BE, TL, and OA. 

\begin{table}[t]
    \centering
    \renewcommand{\arraystretch}{1.2}
    \caption{Quantitative Comparison with Fine-tuning-based methods.}
    \resizebox{1.0\linewidth}{!}{
    \begin{tabular}{c|cccc}
    \hline 
     & MVDream & DreamView & SPAD& SeMv-3D \\
    \hline
    CLIP Score & 30.09 & \textbf{31.57} & 29.42  & \cellcolor{green!10} \underline{30.26} \\
    Mv-Score  & 86.20/100  & 86.37/100 & 87.44/100 & \cellcolor{green!10} \textbf{91.63/100} \\
    \hline
    \end{tabular}
    }
    \label{tab:objective_comparison_ft}
\end{table}

\begin{table}[t]
    \centering
    \renewcommand{\arraystretch}{1.2}
    \caption{User Study of SeMv-3D and Fine-tuning-based methods.}
    \resizebox{1.0\linewidth}{!}{
    \begin{tabular}{c|cccc}
    \hline 
      & MVDream & DreamView & SPAD& SeMv-3D \\
    \hline
    User Prefer & 26.4\%  &22.3\%  & 18.9\%  & \cellcolor{pink!20} \textbf{32.4\%} \\
    Se Consis   & 27.0\%  & 23.6\%  & 18.2\%   & \cellcolor{pink!20} \textbf{31.1\%} \\
    Mv Consis   & 20.3\%  & 16.9\%  & 14.9\%  & \cellcolor{pink!20} \textbf{47.9\%} \\
    \hline
    \end{tabular}
    }
    \label{tab:subjective_comparison_ft}
\end{table}

\begin{figure*}[]
    \centering
    \includegraphics[width=0.9\linewidth]{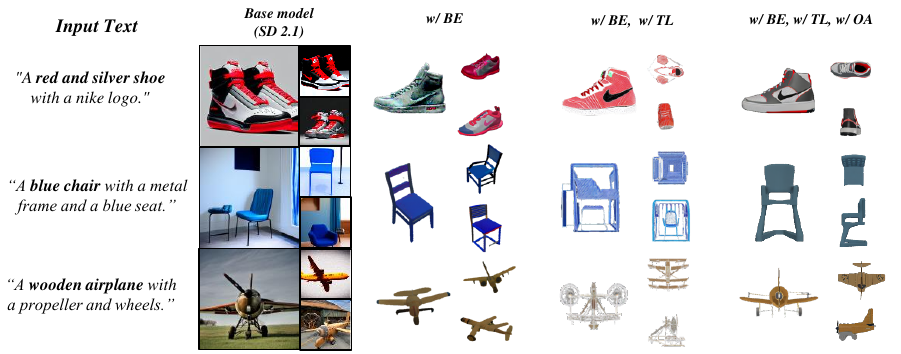}
    \caption{\textbf{Ablation study on the effectiveness of proposed modules in Triplane Prior Learning,} including  Background Elimination (BE) that preserves essential 3D objects without backgrounds, Triplane Learning (TL) that tends to learn the triplane relationships, and Orthogonal Attention (OA) that maintains consistent triplane orthogonal correspondence.}
    \label{fig.ablation_tpl}
\end{figure*}

From the results in Figure \ref{fig.ablation_tpl}, we have the following observations: 
\textbf{(i)} Overall, all proposed components are found to be beneficial, each playing a distinct role in enhancing the spatial consistency and visual quality of the triplane prior.
\textbf{(ii) \textit{Base model.}} Despite producing visually appealing and semantically rich images, the base model suffers from two critical limitations: significant background clutter and a complete lack of cross-plane consistency. The three planes appear as independent images with no coherent 3D structure, making them unsuitable as a consistent prior for 3D generation.
\textbf{(iii) \textit{w/ BE.}} Incorporating the BE module effectively removes extraneous background elements while preserving the semantic integrity of the object. This validates the utility of BE in directing the model's focus toward the object, thereby providing a cleaner foundation for learning spatial correspondences across planes.
\textbf{(iv) \textit{w/ BE, w/ TL.}} The addition of TL introduces explicit triplane-level supervision, which improves spatial coherence to some extent (e.g., the red shoe tends to appear in consistent orthogonal orientations, such as front and side views). However, we observe a noticeable decline in semantic quality and generative fidelity, indicating that naïve supervision alone struggles to balance spatial consistency with high-quality object generation.
\textbf{(v) \textit{w/ BE, w/ TL, w/ OA.}} Integrating OA on top of BE and TL substantially improves the inter-plane spatial alignment while preserving the generative quality of the base model. The triplane images exhibit strong geometric consistency and stable object identity, demonstrating the effectiveness of our orthogonal attention in learning cross-view correspondences.

We further validate the above visual observations through the quantitative results presented in Table~\ref{tab:ablation_tpl}, where we report CLIP Score and Mv-Score across different model configurations.
\textbf{(i)} The base model achieves the highest CLIP Score (30.99), indicating strong semantic alignment and visually appealing image quality. However, its Mv-Score is the lowest (50.28), consistent with its lack of cross-view consistency observed in Figure~\ref{fig.ablation_tpl}. This suggests that, despite high-fidelity appearance, the base model is unsuitable as a coherent prior for 3D generation.
\textbf{(ii)} Incorporating BE leads to a notable improvement in Mv-Score (64.50), while only marginally reducing the CLIP Score (29.31). This confirms that removing background distractions helps the model focus on the object, thereby enhancing spatial correspondence across views.
\textbf{(iii)} With the introduction of TL, Mv-Score further improves to 75.33, but the CLIP Score drops substantially to 24.95. This supports the earlier observation that naïve supervision enhances spatial alignment but may interfere with the preservation of semantic richness and visual quality.
\textbf{(iv)} Finally, integrating OA dramatically boosts Mv-Score to 94.83—the highest among all configurations—while recovering the CLIP Score to 29.67, close to that of the base model. This demonstrates that OA effectively enhances spatial coherence without sacrificing generative fidelity, successfully balancing structure and content.

\subsubsection{\textbf{Effectiveness of the Designed Components in SAT}}  
To evaluate the effectiveness of SAT, we conduct ablation studies by progressively removing its two key components: Cross-Attention (CA) and Orthogonal Attention (OA).

\begin{figure}[t]
\centering
\includegraphics[width=1\linewidth]{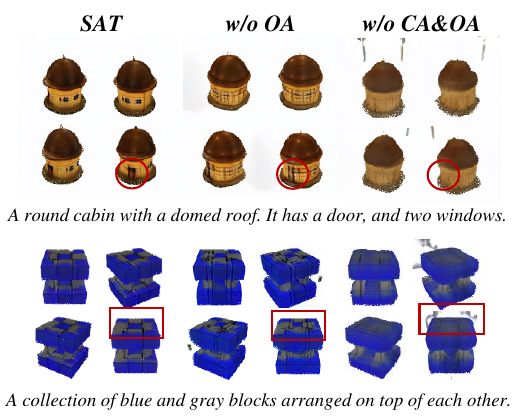}
\caption{\textbf{Ablation study on the effectiveness of CA and OA in SAT.} Cross-attention (CA) facilitates global semantic alignment, while Orthogonal Attention (OA) enhances local semantic consistency. }
\label{fig:ablation_svs}
\end{figure}

\begin{table}[t]
\centering
\small
\caption{\textbf{Quantitative Results of Ablation Study in TPL.} The base model refers to TPL's initialization state, SD-2.1.  BE stands for the Background Elimination module, TL represents the Triplane Learning module, and OA signifies the Orthogonal Attention module. The table shows how each addition affects the Clip Score and Aesthetic Score.}
\begin{tabular}{lcc}
    \toprule
     & CLIP Score & Mv-Score \\
    \midrule
    Base model & \textbf{30.99} & 50.28/100 \\
    w/ BE & 29.31 & 64.50/100 \\
    w/ BE, w/ TL & 24.95 & 75.33/100 \\
    w/ BE, w/ TL, w/ OA & \underline{29.67} & \textbf{94.83/100} \\
    \bottomrule
\end{tabular}
\label{tab:ablation_tpl}
\end{table}
From the visual comparisons in Figure~\ref{fig:ablation_svs}, we make the following observations regarding the impact of each component in SAT:  
\textbf{(i)} The inclusion of both Cross-Attention (CA) and Orthogonal Attention (OA) is critical to achieving high-quality and semantically consistent generation. Their joint effect ensures alignment at both global and local levels.  
\textbf{(ii)} When OA is removed while retaining CA (\textit{w/o OA}), the model largely maintains the overall semantic structure of the object, indicating that CA plays a key role in preserving object-level semantic alignment and suppressing non-semantic artifacts. However, the lack of OA leads to a decline in fine-grained spatial consistency across orthogonal views. For example, the \textit{windows} and \textit{doors} in the top-row cabin example appear vague and underdefined, suggesting that OA is essential for learning accurate local correspondences.  
\textbf{(iii)} Further removing CA (\textit{w/o CA\&OA}) results in a more pronounced degradation. As shown on the right side of Figure~\ref{fig:ablation_svs}, the outputs exhibit severe semantic inconsistencies, structural distortions, and noticeable artifacts, demonstrating that CA is indispensable for maintaining global semantic coherence and suppressing noise.  
\textbf{(iv)} In contrast, the full model equipped with both CA and OA achieves the best results, simultaneously ensuring global semantic fidelity and fine-grained cross-view alignment. These findings validate the effectiveness of jointly incorporating CA and OA to fully realize the potential of the SAT framework.

These observations are further substantiated by the quantitative results in Table~\ref{tab:ablation_svs}.  
\textbf{(i)} The full SAT model yields the highest CLIP Score (30.26) and Mv-Score (91.63), indicating that the combined use of CA and OA leads to accurate semantic grounding and strong multi-view consistency.  
\textbf{(ii)} Removing OA results in a noticeable drop in CLIP Score (27.93) and Mv-Score (87.67), suggesting that although CA helps maintain the global semantic structure, the absence of OA undermines spatial consistency at a finer scale. This aligns with the visual findings in Figure~\ref{fig:ablation_svs}, where local details such as \textit{windows} and \textit{doors} become vague and less coherent.  
\textbf{(iii)} Further removing CA leads to a more pronounced degradation (CLIP Score: 27.52; Mv-Score: 85.06), accompanied by semantic drift, structural distortions, and visible artifacts. This confirms the importance of both components for high-fidelity generation.
Overall, these quantitative results confirm that CA and OA jointly contribute to maintaining semantic coherence and spatial consistency in the SAT framework.

\section{Conclusion}
In this paper, we tackle the fundamental challenge in General Text-to-3D (GT23D) generation—simultaneously ensuring semantic alignment with input text and structural consistency across views. While prior works often prioritize one aspect at the expense of the other, our proposed \textbf{SeMv-3D} framework addresses both in a unified manner.

At its core, SeMv-3D introduces two complementary components: the \textbf{Triplane Prior Learning (TPL)} module, which constructs a geometrically consistent triplane prior by capturing spatial correspondences across orthogonal planes via Orthogonal Attention; and the \textbf{Prior-based Semantic Aligning in Triplanes (SAT)} module, which reinforces semantic consistency by aligning triplane features with textual guidance at both global and local levels.

\begin{table}[t]
\centering
\small
\caption{\textbf{Quantitative Results of Ablation Study in SAT.}  OA stands for the Orthogonal Attention, CA represents the Cross Attention. The table shows how each removal affects the semantic and multi-view consistency.}
\begin{tabular}{lccc}
    \toprule
     & SAT & w/o OA & w/o OA\&CA \\
    \midrule
   CLIP Score  & \textbf{30.26} & 27.93 &  27.52\\
    Mv-Score & \textbf{91.63/100} & 87.67/100 & 85.06/100 \\
    \bottomrule
\end{tabular}
\label{tab:ablation_svs}
\end{table}

Comprehensive experiments demonstrate that SeMv-3D achieves state-of-the-art performance in multi-view consistency while maintaining strong semantic fidelity. These results underscore the effectiveness of jointly modeling geometry and semantics within a unified triplane-based framework, and suggest SeMv-3D as a promising foundation for future research in general-purpose 3D and even 4D generation.


\bibliography{reference}
\bibliographystyle{IEEEtran}

\vfill

\end{document}